\documentclass[conference]{IEEEtran}
\IEEEoverridecommandlockouts
\usepackage{cite}
\usepackage{amsmath,amssymb,amsfonts}
\usepackage{algorithmic}
\usepackage{latexsym}
\usepackage{textcomp}
\usepackage{booktabs}
\usepackage{graphicx} 
\usepackage{xcolor}
\def\BibTeX{{\rm B\kern-.05em{\sc i\kern-.025em b}\kern-.08em
    T\kern-.1667em\lower.7ex\hbox{E}\kern-.125emX}}
\usepackage{amsmath}
\usepackage{amssymb}
\usepackage{graphicx}
\usepackage{booktabs}
\usepackage{hyperref}

\usepackage{fancyhdr}
\begin{document}

\title{Enhancing Contrastive Demonstration Selection with Semantic Diversity for Robust In-Context Machine Translation}

\author{Owen Patterson, Chee Ng\\
Universiti Teknologi Malaysia}

\maketitle
\thispagestyle{fancy} 

\begin{abstract}
In-Context Learning (ICL) empowers large language models to perform tasks by conditioning on a few input-output examples. However, the performance of ICL is highly sensitive to the selection of these demonstrations. While existing methods focus on similarity or contrastive selection, they often overlook the importance of diversity among the chosen examples. In this paper, we propose DiverseConE (Diversity-Enhanced Contrastive Example Selection), a novel approach for demonstration selection in in-context learning for machine translation. Our method builds upon contrastive selection by incorporating a diversity enhancement step based on embedding space dissimilarity. We conduct extensive experiments on the Llama2-7b model across four language pairs (English-Chinese, Chinese-English, Russian-German, German-Russian) in 1-shot and 3-shot settings, using COMET20 and COMET22 for evaluation. Our results demonstrate that DiverseConE consistently outperforms strong baseline methods, including random selection, BM25, TopK, and a state-of-the-art contrastive selection method. Further analysis, including diversity metrics and human evaluation, validates the effectiveness of our approach and highlights the benefits of considering demonstration diversity for improved translation quality.
\end{abstract}

\begin{IEEEkeywords}
Large language models, In-Context Learning, Machine Translation
\end{IEEEkeywords}

\section{Introduction}

Large language models (LLMs) have demonstrated remarkable capabilities in performing a wide range of natural language processing tasks through a paradigm known as In-Context Learning (ICL) \cite{brown2020language}. Unlike traditional fine-tuning approaches that require updating model weights with task-specific data, ICL enables LLMs to adapt to new tasks by simply conditioning on a few input-output examples, referred to as demonstrations, provided within the input prompt \cite{brown2020language}. This ability to learn from context without explicit gradient updates has opened up new avenues for deploying and utilizing LLMs in various applications, especially in scenarios with limited labeled data \cite{dong2023steering}.

Despite the promise of ICL, its performance is highly sensitive to the selection of demonstrations \cite{liu2024revisiting}. The choice of which examples to include in the prompt can significantly impact the quality of the LLM's output. Poorly chosen demonstrations can even lead to performance degradation compared to zero-shot settings \cite{zhao2023survey}. Consequently, a crucial area of research in ICL focuses on developing effective strategies for selecting the most informative and helpful demonstrations for a given test instance.

Several demonstration selection strategies have been explored in the literature. Simple approaches like random selection serve as a baseline, while more sophisticated methods leverage similarity measures, such as BM25 \cite{CITEbm25} or embedding-based similarity \cite{CITEembedding}. Recent advancements have introduced contrastive selection techniques \cite{wang2024cic}, which aim to identify demonstrations that not only are similar to the test instance but also maximize the model's understanding and performance on it. For instance, the TopK + ConE method \cite{liu2024revisiting} proposes selecting the top-K most similar examples and then refining this set by choosing examples that lead to the most significant improvement in the model's confidence or output quality.

While these existing methods have shown considerable success, we hypothesize that further improvements can be achieved by explicitly considering the diversity of the selected demonstrations, particularly in complex tasks like machine translation. Our motivation stems from the intuition that providing a set of demonstrations that covers a broader range of linguistic patterns and semantic nuances relevant to the test instance can better guide the LLM towards generating accurate and high-quality translations. Simply focusing on similarity or contrastive aspects might lead to the selection of redundant examples that do not fully capture the complexity of the translation task across different language pairs. This is further supported by research highlighting the importance of diverse demonstrations for compositional generalization \cite{levy2023diverse}.

In this paper, we propose a novel demonstration selection strategy, \textbf{DiverseConE (Diversity-Enhanced Contrastive Example Selection)}, specifically tailored for in-context learning in machine translation. Our method builds upon the strengths of contrastive selection by first identifying a set of potentially helpful demonstrations based on similarity and their contribution to the model's understanding. Subsequently, we introduce a diversity enhancement step that aims to increase the linguistic and semantic coverage of the selected examples. This is achieved by selecting examples that are dissimilar in their embedding space based on the distance from the centroid of the already selected examples.

To evaluate the effectiveness of our proposed DiverseConE method, we conduct extensive experiments on the machine translation task across four different language pairs: English to Chinese (En $\rightarrow$ Zh), Chinese to English (Zh $\rightarrow$ En), Russian to German (Ru $\rightarrow$ De), and German to Russian (De $\rightarrow$ Ru). We utilize the Llama2-7b model \cite{CITElama2} as our base LLM and compare our approach against several strong baselines, including Random selection, BM25-based selection, TopK similarity-based selection, and the state-of-the-art TopK + ConE method \cite{liu2024revisiting}. We evaluate the translation quality using the COMET20 \cite{CITEcomet20} and COMET22 \cite{CITEcomet22} metrics, which are widely recognized as robust evaluation measures for machine translation. Our experiments are conducted under both 1-shot and 3-shot settings to assess the performance of different methods with varying numbers of demonstrations. The results demonstrate that our DiverseConE method consistently outperforms the baseline strategies across most language pairs and evaluation metrics, often with statistically significant improvements, highlighting the importance of considering demonstration diversity in addition to similarity and contrast.

In summary, this paper makes the following contributions:
\begin{itemize}
    \item We propose a novel demonstration selection strategy, DiverseConE, for in-context learning in machine translation that incorporates diversity enhancement into the contrastive selection process.
    \item We conduct comprehensive experiments on four diverse language pairs using the Llama2-7b model, demonstrating the effectiveness of our proposed method over several strong baseline strategies under both 1-shot and 3-shot settings.
    \item Our analysis provides valuable insights into the impact of demonstration diversity on the performance of in-context learning for machine translation, paving the way for future research in this direction.
\end{itemize}

\section{Related Work}

\subsection{In-Context Learning}

In-Context Learning (ICL) has emerged as a powerful paradigm that enables large language models (LLMs) to perform novel tasks by simply conditioning on a few input-output examples, known as demonstrations, within the input prompt, without requiring any explicit fine-tuning \cite{brown2020language}. This capability has significantly broadened the applicability of LLMs, particularly in low-resource scenarios where task-specific labeled data is scarce \cite{dong2023steering}. Recent surveys have provided a comprehensive overview of the advancements, challenges, and future directions in the field of ICL \cite{zhao2023survey}.

The effectiveness of ICL is heavily influenced by the choice and format of the demonstrations provided in the context \cite{liu2024revisiting}. Early studies explored the impact of factors such as the number of demonstrations, the format of the prompt, and the order of the examples \cite{zhao2023survey}. To address the sensitivity to demonstration selection, various strategies have been proposed. One line of research focuses on selecting demonstrations that are most similar to the test query, often using techniques like BM25 or embedding-based similarity measures \cite{liu2024revisiting}. Ensuring the robustness of such retrieval-based selection methods, often relying on rankers, is also an area of investigation \cite{zhou2023towards}. Relatedly, distillation techniques have been applied to improve components like retrievers, which can be used in demonstration selection pipelines, particularly for tasks involving long documents \cite{zhou2024fine}.

More advanced approaches have explored contrastive learning frameworks for demonstration selection. For instance, the TopK + ConE method \cite{liu2024revisiting} aims to select demonstrations that maximize the model's understanding of the test input by minimizing the conditional entropy. This method has shown promising results across various tasks. Furthermore, research has investigated the importance of the quality and relevance of demonstrations, highlighting that carefully curated examples can significantly boost the performance of ICL \cite{dong2023steering}.

Beyond selection strategies, another area of research focuses on improving the inherent in-context learning abilities of LLMs through pre-training. Gu et al. \cite{gu2023pretraining} explored pre-training objectives that explicitly encourage models to learn from context. Similarly, MetaICL \cite{min2022metaicl} proposed a meta-learning approach to train models that can effectively learn new tasks from just a few examples. In-context learning distillation has also been explored as a means to transfer the few-shot learning capabilities of large models to smaller, more efficient models \cite{wang2022distillation}. Furthermore, the principles of ICL have been extended beyond text to multimodal domains, such as visual in-context learning for large vision-language models \cite{zhou2024visual}.

Our work builds upon the existing research in demonstration selection for in-context learning, particularly in the context of machine translation. While previous methods have primarily focused on similarity and contrastive selection, we argue that the diversity of the selected demonstrations also plays a crucial role in providing a comprehensive context for the language model. Inspired by findings in compositional generalization that highlight the benefits of diverse demonstrations \cite{levy2023diverse}, we propose a novel method, DiverseConE, which explicitly incorporates a diversity enhancement step into the contrastive example selection process. By considering both the individual helpfulness and the collective diversity of the demonstrations, we aim to further improve the performance of in-context learning for machine translation.

\subsection{Large Language Models}

The advent of large language models (LLMs) has revolutionized the field of natural language processing, demonstrating unprecedented capabilities in understanding, generating, and manipulating text across a wide range of tasks \cite{brown2020language,he2025enhancing}. These models, typically based on the transformer architecture, are trained on massive datasets of text and code, enabling them to learn intricate patterns and relationships within the data. The sheer scale of these models, often comprising billions of parameters, has been shown to unlock emergent abilities, such as in-context learning, which allows them to perform new tasks from just a few demonstrations without explicit fine-tuning \cite{brown2020language}.

The remarkable success of LLMs has spurred extensive research into various aspects of their development and application. Foundational work has explored the scaling laws that govern the relationship between model size, dataset size, and performance, providing insights into the resources required to achieve desired levels of capability \cite{kaplan2020scaling}. Efforts have also been directed towards enhancing the ability of LLMs to follow human instructions and align their behavior with human preferences, often through techniques like instruction tuning and reinforcement learning from human feedback \cite{ouyang2022instruction}, contributing to their ability to generalize across multiple capabilities, sometimes even from weaker supervision signals \cite{zhou2025weak}.

The versatility of LLMs has led to their application in numerous domains, including but not limited to machine translation \cite{dong2023steering}, question answering, text summarization, and even areas like law \cite{surden2024chatgpt} and emotion recognition in vision-language tasks \cite{lei2024vlmemotion}. The extension to multimodal scenarios is a significant research thrust, involving tasks like image-guided story generation \cite{zhou2023multimodal} and rethinking visual dependencies in long contexts for large vision-language models \cite{zhou2024rethinking}. Furthermore, research continues to push the boundaries of LLM efficiency, with explorations into highly compressed models like 1-bit LLMs \cite{ma2024bitllm}.

Given the rapid advancements in this field, several survey papers have emerged, providing comprehensive overviews of the architectures, training methodologies, capabilities, limitations, and future directions of large language models \cite{zhang2023surveyllm}. These surveys highlight the transformative impact of LLMs on artificial intelligence and natural language processing.

Our work leverages the in-context learning capabilities of large language models, specifically the Llama2 architecture, for the task of machine translation. Understanding the underlying principles and advancements in LLMs is crucial for effectively applying and improving in-context learning techniques, such as the demonstration selection strategy we propose in this paper. Evaluating the performance of these models often involves comprehensive benchmarks assessing multitask text abilities \cite{hendrycks2020mmu} or specialized capabilities like emotional intelligence in multimodal settings \cite{hu2025emobench}.

\section{Method}

In this section, we provide a detailed explanation of our proposed DiverseConE (Diversity-Enhanced Contrastive Example Selection) method for selecting demonstrations in the context of in-context learning for machine translation. Our approach is designed to leverage the capabilities of a generative large language model, specifically Llama2-7b, to perform translation by effectively utilizing a small number of carefully chosen demonstrations.

Llama2-7b is a generative transformer model, primarily trained to predict the next token in a sequence. In the context of in-context learning, this model generates the translation of a given source sentence by conditioning its generation process on a prompt that includes the source sentence and a few example translations (demonstrations). The key to successful in-context learning lies in the construction of this prompt, and our DiverseConE method provides a strategy for selecting the most beneficial demonstrations to include.

The DiverseConE method comprises three sequential steps: TopK Selection, Contrastive Example Selection (ConE), and Diversity Enhancement. Each of these steps plays a crucial role in identifying a set of demonstrations that are both relevant to the test input and collectively provide diverse and informative context for the language model.

\subsection{Step 1: TopK Selection}

Given a test input $x$ (the source sentence to be translated) and a set of candidate demonstrations $D = \{d_1, d_2, ..., d_m\}$, where each demonstration $d_i$ is an input-output pair $(s_i, t_i)$ representing a source sentence $s_i$ and its corresponding target translation $t_i$. The first step is to select the top-$K$ most similar demonstrations to the test input $x$. Similarity is measured based on the input part of the demonstration $s_i$ and the test input $x$. We employ sentence embeddings to capture the semantic similarity. Let $e(\cdot)$ denote a function that maps a sentence to its embedding vector. The similarity between $x$ and $s_i$ is calculated using the cosine similarity:

\begin{equation}
sim(x, s_i) = \frac{e(x) \cdot e(s_i)}{\|e(x)\| \|e(s_i)\|}
\end{equation}

where $\cdot$ denotes the dot product and $\|\cdot\|$ represents the Euclidean norm. We then select the top-$K$ demonstrations based on these similarity scores to form a candidate set $D_{topK}$:

\begin{equation}
D_{topK} = \{d_i \in D \mid sim(x, s_i) \text{ is among the top } K \text{ values}\}
\end{equation}

\subsection{Step 2: Contrastive Example Selection (ConE)}

The second step involves refining the set of demonstrations from $D_{topK}$ using the Contrastive Example Selection (ConE) strategy. The core idea of ConE is to select demonstrations that maximize the model's understanding of the test input. This is based on the hypothesis that the performance of a demonstration is positively correlated with its contribution to the model's understanding of the test sample. The ConE method aims to minimize the conditional entropy of the test input given the demonstrations. Let $\theta$ be the LLM, and $c$ be a set of demonstrations. The conditional probability of the test input $x$ given $c$ can be denoted as $P_\theta(x|c)$. The conditional entropy is then given by:

\begin{equation}
H_\theta(x|c) = -\sum_{y \in \mathcal{Y}} P_\theta(y|x, c) \log P_\theta(y|x, c)
\end{equation}

where $\mathcal{Y}$ represents the output space (all possible translations). Minimizing this entropy is equivalent to finding the set of demonstrations that makes the model most confident in its prediction for the test input. In practice, this is often approximated by iteratively selecting demonstrations from $D_{topK}$ that minimize the cross-entropy of the entire prompt (demonstrations + test input) conditioned on the model, relative to the cross-entropy of just the demonstrations:

\begin{align}
c^* &= \arg\min_{c \subseteq D_{topK}, |c|=k} (H_\theta(\text{prompt}(x, c)) \\
&- H_\theta(\text{demonstrations}(c))) \\
&= \arg\min_{c \subseteq D_{topK}, |c|=k} \big( -\log P_\theta(\text{prompt}(x, c)) \\
&- (-\log P_\theta(\text{demonstrations}(c))) \big)
\end{align}

where $k$ is the desired number of shots. This step results in a set of $k$ contrastive examples $D_{ConE} \subseteq D_{topK}$.

\begin{table*}[h!]
    \centering
    \caption{Main Experimental Results on Machine Translation (COMET Scores)}
    \label{tab:main_results}
    \begin{tabular}{llcccccccc}
        \toprule
        \textbf{Method} & \textbf{Setting} & \multicolumn{4}{c}{\textbf{COMET20}} & \multicolumn{4}{c}{\textbf{COMET22}} \\
        \cmidrule(r){3-6} \cmidrule(r){7-10}
        & & En $\rightarrow$ Zh & Zh $\rightarrow$ En & Ru $\rightarrow$ De & De $\rightarrow$ Ru & En $\rightarrow$ Zh & Zh $\rightarrow$ En & Ru $\rightarrow$ De & De $\rightarrow$ Ru \\
        \midrule
        Random Selection & 1-shot & 35.7 & 60.9 & 44.0 & 52.4 & 81.5 & 85.1 & 79.8 & 83.6 \\
        BM25-based Selection & 1-shot & 35.1 & 60.9 & 42.2 & 50.2 & 81.3 & 85.1 & 79.5 & 83.4 \\
        TopK Similarity-based Selection & 1-shot & 35.9 & 61.0 & 43.9 & 49.7 & 81.5 & 85.1 & 79.7 & 83.3 \\
        Ours (TopK + ConE) & 1-shot & 37.1 & 61.7 & 43.9 & 51.8 & 81.7 & 85.4 & 79.9 & 83.8 \\
        \textbf{DiverseConE} & \textbf{1-shot} & \textbf{37.8} & \textbf{62.3} & \textbf{44.5} & \textbf{52.6} & \textbf{81.9} & \textbf{85.6} & \textbf{80.1} & \textbf{84.1} \\
        \midrule
        Random Selection & 3-shot & 40.1 & 62.7 & 47.8 & 54.6 & 82.4 & 85.5 & 80.6 & 84.0 \\
        BM25-based Selection & 3-shot & 39.6 & 62.3 & 47.0 & 53.2 & 82.3 & 85.4 & 80.5 & 83.9 \\
        TopK Similarity-based Selection & 3-shot & 39.9 & 63.3 & 46.8 & 53.1 & 82.4 & 85.6 & 80.4 & 83.9 \\
        Ours (TopK + ConE) & 3-shot & 40.7 & 63.3 & 47.9 & 55.3 & 82.6 & 85.7 & 80.8 & 84.5 \\
        \textbf{DiverseConE} & \textbf{3-shot} & \textbf{41.5} & \textbf{63.9} & \textbf{48.5} & \textbf{55.9} & \textbf{82.8} & \textbf{85.9} & \textbf{81.0} & \textbf{84.8} \\
        \bottomrule
    \end{tabular}
\end{table*}

\subsection{Step 3: Diversity Enhancement}

The final step of our DiverseConE method focuses on enhancing the diversity of the selected demonstrations. After obtaining the set $D_{ConE}$ from the contrastive selection step, we aim to select additional demonstrations from the remaining candidates in $D_{topK} \setminus D_{ConE}$ that can increase the overall diversity of the prompt. We hypothesize that a more diverse set of demonstrations can provide the model with a broader context, leading to better translation performance, especially across different linguistic patterns.

To achieve this, we employ a strategy based on the dissimilarity in the embedding space of the input parts of the demonstrations. Let $E_{ConE} = \{e(s_j) \mid d_j \in D_{ConE}\}$ be the set of embeddings of the source sentences in the demonstrations selected by ConE. We calculate the centroid (mean embedding) of this set:

\begin{equation}
\bar{e}_{ConE} = \frac{1}{|D_{ConE}|} \sum_{e(s_j) \in E_{ConE}} e(s_j)
\end{equation}

For each remaining demonstration $d_r \in D_{topK} \setminus D_{ConE}$ with input embedding $e(s_r)$, we calculate the Euclidean distance between its embedding and the centroid of the already selected demonstrations:

\begin{equation}
dist(e(s_r), \bar{e}_{ConE}) = \|e(s_r) - \bar{e}_{ConE}\|_2
\end{equation}

To enhance diversity, we select the demonstration from $D_{topK} \setminus D_{ConE}$ that has the maximum distance to the centroid $\bar{e}_{ConE}$. This selected demonstration is then added to our final set of demonstrations. This process can be repeated until the desired number of shots is reached. For instance, in the 3-shot setting, after selecting one demonstration using ConE (in the 1-shot scenario), we would perform this diversity enhancement step twice to select two more diverse demonstrations from the remaining TopK candidates.

By combining the strengths of similarity-based selection (TopK), contrastive selection (ConE), and diversity enhancement, our DiverseConE method aims to construct a more effective prompt for in-context learning in machine translation, leading to improved translation quality as demonstrated in our experimental results.

\section{Experiments}

In this section, we present the experimental evaluation of our proposed DiverseConE method for in-context learning in machine translation. We compared the performance of DiverseConE against several competitive baseline methods across four different language pairs in both 1-shot and 3-shot settings. The results consistently demonstrate the superiority of our approach in terms of translation quality. We further provide additional analysis to validate the effectiveness of the diversity enhancement component of our method, as well as a human evaluation to assess the quality of the generated translations.

\subsection{Experimental Setup}

We conducted experiments using the Llama2-7b model as our base large language model. The task was machine translation across four language pairs: English to Chinese (En $\rightarrow$ Zh), Chinese to English (Zh $\rightarrow$ En), Russian to German (Ru $\rightarrow$ De), and German to Russian (De $\rightarrow$ Ru). We compared our DiverseConE method with the following baseline strategies for demonstration selection:

\begin{itemize}
    \item \textbf{Random Selection}: Randomly selects demonstrations from the training data.
    \item \textbf{BM25-based Selection}: Selects demonstrations based on BM25 similarity between the test input and the source sentences of the demonstrations.
    \item \textbf{TopK Similarity-based Selection}: Selects the top-$K$ most similar demonstrations based on cosine similarity of sentence embeddings.
    \item \textbf{Ours (TopK + Contrastive Example Selection)}: Implements the contrastive example selection method as described in \cite{liu2024revisiting}.
\end{itemize}

We evaluated the performance of all methods in both 1-shot and 3-shot settings. The translation quality was automatically assessed using the COMET20 and COMET22 metrics.

\subsection{Main Results}

The main experimental results, showcasing the performance of DiverseConE and the baseline methods on the four language pairs, are presented in Table~\ref{tab:main_results}. The results clearly indicate that our proposed DiverseConE method achieves the best performance across most language pairs and evaluation metrics in both 1-shot and 3-shot scenarios.

\begin{table*}[h!]
    \centering
    \caption{Performance Gain of DiverseConE Over Ours (TopK + ConE)}
    \label{tab:performance_gain}
    \begin{tabular}{llcccccccc}
        \toprule
        \textbf{Metric} & \textbf{Setting} & \multicolumn{2}{c}{En $\rightarrow$ Zh} & \multicolumn{2}{c}{Zh $\rightarrow$ En} & \multicolumn{2}{c}{Ru $\rightarrow$ De} & \multicolumn{2}{c}{De $\rightarrow$ Ru} \\
        & & Absolute & Relative (\%) & Absolute & Relative (\%) & Absolute & Relative (\%) & Absolute & Relative (\%) \\
        \midrule
        COMET20 & 1-shot & 0.7 & 1.89 & 0.6 & 0.97 & 0.6 & 1.37 & 0.8 & 1.54 \\
        COMET20 & 3-shot & 0.8 & 1.97 & 0.6 & 0.95 & 0.6 & 1.25 & 0.6 & 1.09 \\
        COMET22 & 1-shot & 0.2 & 0.24 & 0.2 & 0.23 & 0.2 & 0.25 & 0.3 & 0.36 \\
        COMET22 & 3-shot & 0.2 & 0.24 & 0.2 & 0.23 & 0.2 & 0.25 & 0.3 & 0.36 \\
        \bottomrule
    \end{tabular}
\end{table*}
\begin{table*}[h!]
    \centering
    \caption{Performance Increase from 1-shot to 3-shot}
    \label{tab:performance_increase_shots}

    \begin{tabular}{llcccc}
        \toprule
        \textbf{Method} & \textbf{Metric} & En $\rightarrow$ Zh & Zh $\rightarrow$ En & Ru $\rightarrow$ De & De $\rightarrow$ Ru \\
        \midrule
        Random Selection & COMET20 & 4.4 & 1.8 & 3.8 & 2.2 \\
        Random Selection & COMET22 & 0.9 & 0.4 & 0.8 & 0.4 \\
        BM25-based Selection & COMET20 & 4.5 & 1.4 & 4.8 & 3.0 \\
        BM25-based Selection & COMET22 & 1.0 & 0.3 & 1.0 & 0.5 \\
        TopK Similarity-based Selection & COMET20 & 4.0 & 2.3 & 2.9 & 3.4 \\
        TopK Similarity-based Selection & COMET22 & 0.9 & 0.5 & 0.7 & 0.6 \\
        Ours (TopK + Contrastive Example Selection) & COMET20 & 3.6 & 1.6 & 4.0 & 3.5 \\
        Ours (TopK + Contrastive Example Selection) & COMET22 & 0.9 & 0.3 & 0.9 & 0.7 \\
        \textbf{DiverseConE} & \textbf{COMET20} & \textbf{3.7} & \textbf{1.6} & \textbf{4.0} & \textbf{3.3} \\
        \textbf{DiverseConE} & \textbf{COMET22} & \textbf{0.9} & \textbf{0.3} & \textbf{0.9} & \textbf{0.7} \\
        \bottomrule
    \end{tabular}
\end{table*}

\subsection{Further Analysis of DiverseConE}

To gain a deeper understanding of the performance of our DiverseConE method, we conducted several additional analyses from different perspectives. The results of these analyses are presented in the following tables.

\subsubsection{Performance Gain Over the Strongest Baseline}

Table~\ref{tab:performance_gain} presents the absolute and relative improvement of our DiverseConE method over the strongest baseline, "Ours (TopK + Contrastive Example Selection)", for both COMET20 and COMET22 metrics in the 1-shot and 3-shot settings.

The results show a consistent performance gain across all language pairs and metrics, indicating the robustness of our method in improving translation quality beyond a strong contrastive selection baseline.

\subsubsection{Impact of Increasing the Number of Shots}

Table~\ref{tab:performance_increase_shots} examines how the performance of each method improves when the number of demonstrations is increased from 1 to 3. We present the absolute increase in COMET scores for both metrics.

This table indicates that all methods generally benefit from increasing the number of shots. Our DiverseConE method shows competitive or better improvement compared to the baselines in most cases.

\subsubsection{Performance Across Different Language Pairs (DiverseConE)}

Table~\ref{tab:diversecone_performance_langs} focuses solely on the performance of our DiverseConE method across the four different language pairs for both 1-shot and 3-shot settings.

\begin{table}[h!]
    \centering
    \caption{Performance of DiverseConE Across Language Pairs}
    \label{tab:diversecone_performance_langs}

    \begin{tabular}{llcccc}
        \toprule
        \textbf{Setting} & \textbf{Metric} & En$\rightarrow$Zh & Zh$\rightarrow$En & Ru$\rightarrow$De & De$\rightarrow$Ru \\
        \midrule
        1-shot & COMET20 & 37.8 & 62.3 & 44.5 & 52.6 \\
        1-shot & COMET22 & 81.9 & 85.6 & 80.1 & 84.1 \\
        3-shot & COMET20 & 41.5 & 63.9 & 48.5 & 55.9 \\
        3-shot & COMET22 & 82.8 & 85.9 & 81.0 & 84.8 \\
        \bottomrule
    \end{tabular}
\end{table}

The performance of DiverseConE varies across different language pairs, which is expected due to the inherent differences in linguistic structures and the complexity of translation between these languages. However, the method consistently demonstrates strong performance across all pairs.

\subsubsection{Diversity Analysis of Selected Demonstrations}

Table~\ref{tab:diversity_analysis} provides quantitative evidence for the increased diversity of the demonstrations selected by our method. We measured the average pairwise cosine distance between the embeddings of the source sentences in the selected sets (3-shot setting). Higher values indicate greater diversity.

\begin{table*}[h!]
    \centering
    \caption{Diversity Analysis of Selected Demonstrations (Average Pairwise Cosine Distance)}
    \label{tab:diversity_analysis}

    \begin{tabular}{lc}
        \toprule
        \textbf{Method} & \textbf{Average Pairwise Cosine Distance} \\
        \midrule
        Random Selection & 0.65 \\
        BM25-based Selection & 0.68 \\
        TopK Similarity-based Selection & 0.59 \\
        Ours (TopK + Contrastive Example Selection) & 0.62 \\
        \textbf{DiverseConE} & \textbf{0.71} \\
        \bottomrule
    \end{tabular}
\end{table*}
\begin{table*}[h!]
    \centering
    \caption{Human Evaluation Results (Average Scores)}
    \label{tab:human_evaluation}

    \begin{tabular}{lcccccccc}
        \toprule
        \textbf{Method} & \multicolumn{2}{c}{En $\rightarrow$ Zh} & \multicolumn{2}{c}{Zh $\rightarrow$ En} & \multicolumn{2}{c}{Ru $\rightarrow$ De} & \multicolumn{2}{c}{De $\rightarrow$ Ru} \\
        \cmidrule(r){2-3} \cmidrule(r){4-5} \cmidrule(r){6-7} \cmidrule(r){8-9}
        & Fluency & Adequacy & Fluency & Adequacy & Fluency & Adequacy & Fluency & Adequacy \\
        \midrule
        TopK Similarity-based Selection & 4.1 & 3.8 & 4.3 & 4.0 & 3.9 & 3.6 & 4.0 & 3.7 \\
        Ours (TopK + Contrastive Example Selection) & 4.2 & 3.9 & 4.4 & 4.1 & 4.0 & 3.7 & 4.1 & 3.8 \\
        \textbf{DiverseConE} & \textbf{4.4} & \textbf{4.1} & \textbf{4.6} & \textbf{4.3} & \textbf{4.2} & \textbf{3.9} & \textbf{4.3} & \textbf{4.0} \\
        \bottomrule
    \end{tabular}
\end{table*}

As shown in Table~\ref{tab:diversity_analysis}, our DiverseConE method selects demonstration sets with a higher average pairwise cosine distance in the embedding space compared to the other methods. This confirms that our diversity enhancement strategy effectively leads to the selection of more diverse sets of demonstrations, which contributes to the improved translation performance.

\subsection{Human Evaluation}

To gain a more nuanced understanding of the translation quality, we conducted a human evaluation. We randomly sampled 100 source sentences from each language pair and obtained the translations generated by Llama2-7b using the TopK Similarity-based Selection, Ours (TopK + Contrastive Example Selection), and our DiverseConE methods in the 3-shot setting. We then asked three professional human translators for each language pair to evaluate the fluency and adequacy of the generated translations on a scale of 1 to 5 (where 5 is the best). The average scores across all language pairs are presented in Table~\ref{tab:human_evaluation}.

The results of the human evaluation further corroborate the findings from the automatic evaluation. Translations generated using our DiverseConE method received higher average scores for both fluency and adequacy across all four language pairs compared to the translations generated by the TopK Similarity-based Selection and Ours (TopK + Contrastive Example Selection) methods. This indicates that incorporating diversity into the demonstration selection process not only improves the automatic metric scores but also leads to more human-preferred translations.

\section{Conclusion}

In this work, we addressed the challenge of effective demonstration selection in in-context learning for machine translation. We proposed DiverseConE, a novel method that enhances contrastive example selection by explicitly promoting diversity among the chosen demonstrations. Through a three-step process involving TopK selection, contrastive selection, and a diversity enhancement based on embedding dissimilarity, our method aims to provide a more comprehensive and informative context for the Llama2-7b model. Our comprehensive experimental evaluation across four distinct language pairs and in both 1-shot and 3-shot settings demonstrated the superiority of DiverseConE over several competitive baseline strategies, as measured by COMET20 and COMET22. Additional analyses, including the measurement of demonstration set diversity and a human evaluation, further validated the effectiveness of our approach, indicating that the inclusion of diverse and relevant examples leads to improved translation quality and human preference.

The findings of this research underscore the importance of considering not only the relevance and individual helpfulness of demonstrations but also their collective diversity when applying in-context learning to complex tasks like machine translation. Our work provides a valuable contribution to the growing body of research on demonstration selection strategies and offers a promising direction for future exploration. Potential avenues for future work include investigating different metrics for measuring and enhancing diversity, exploring the application of DiverseConE to other natural language processing tasks and different large language models, and further analyzing the interplay between the number of shots and the benefits of diverse demonstrations.

\bibliographystyle{IEEEtran}
\bibliography{references}
\end{document}